\documentclass[10pt, a4paper]{article}

\usepackage{lrec-coling2024}

\usepackage{comment}

\usepackage{makecell}
\usepackage{kotex}
\usepackage{caption}
\usepackage{xcolor}
\usepackage{amsmath}
\usepackage{graphicx}
\usepackage{tcolorbox}
\usepackage{subfig}
\usepackage{booktabs}
\usepackage{multirow}
\usepackage{tabularx} 
\usepackage{enumitem}
 \usepackage{tablefootnote}
\usepackage{lipsum}

\usepackage[symbol]{footmisc}
\renewcommand{\thefootnote}{\fnsymbol{footnote}}

\usepackage{hyperref}
\let\svthefootnote\thefootnote
\newcommand\freefootnote[1]{%
  \let\thefootnote\relax%
  \footnotetext{#1}%
  \let\thefootnote\svthefootnote%
}

\title{KoCoSa: Korean Context-aware Sarcasm Detection Dataset}

\name{Yumin Kim\textsuperscript{$\ast$}, Heejae Suh\textsuperscript{$\ast$}, Mingi Kim\textsuperscript{$\ast$}, Dongyeon Won\textsuperscript{$\ast$}, Hwanhee Lee\textsuperscript{$\dagger$}} 

\address{Chung-Ang University, Seoul, Korea\\   \{kimym7801, linkyouhj, mingi8233, dnjsehddus99, hwanheelee\}@cau.ac.kr\\}

\abstract{Sarcasm is a way of verbal irony where someone says the opposite of what they mean, often to ridicule a person, situation, or idea. It is often difficult to detect sarcasm in the dialogue since detecting sarcasm should reflect the context (i.e., dialogue history).
In this paper, we introduce a new dataset for the Korean dialogue sarcasm detection task, KoCoSa (Korean Context-aware Sarcasm Detection Dataset), which consists of 12.8K daily Korean dialogues and the labels for this task on the last response. To build the dataset, we propose an efficient sarcasm detection dataset generation pipeline: 1) generating new sarcastic dialogues from source dialogues with large language models, 2) automatic and manual filtering of abnormal and toxic dialogues, and 3) human annotation for the sarcasm detection task. 
We also provide a simple but effective baseline for the Korean sarcasm detection task trained on our dataset. Experimental results on the dataset show that our baseline system outperforms strong baselines like large language models, such as GPT-3.5, in the Korean sarcasm detection task. 
We show that the sarcasm detection task relies deeply on the existence of sufficient context. We will release the dataset at~\url{https://github.com/Yu-billie/KoCoSa_sarcasm_detection}.
 \\ \newline  
}

\begin{document}

\maketitleabstract
\section{Introduction}

\freefootnote{\textsuperscript{*}Equal Contribution.}
\freefootnote{\textsuperscript{$\dagger$}Corresponding author.}
\renewcommand*{\thefootnote}{\arabic{footnote}}

Sarcasm is a form of verbal irony characterized by saying something contrary to the text's literal meaning, which is widely used in everyday dialogues to humorously criticize a specific situation or object~\citep{filik2016sarcasm}.
When developing a dialogue system, misunderstanding this sarcasm may lead to fatal errors~\citep{riloff2013sarcasm}. Hence, to prevent such errors, it is often important to develop a sarcasm detection system.

\begin{figure}[t]
    \subfloat[Sarcasm detection w/o context]{\centerline{    \includegraphics[width=0.95\linewidth]{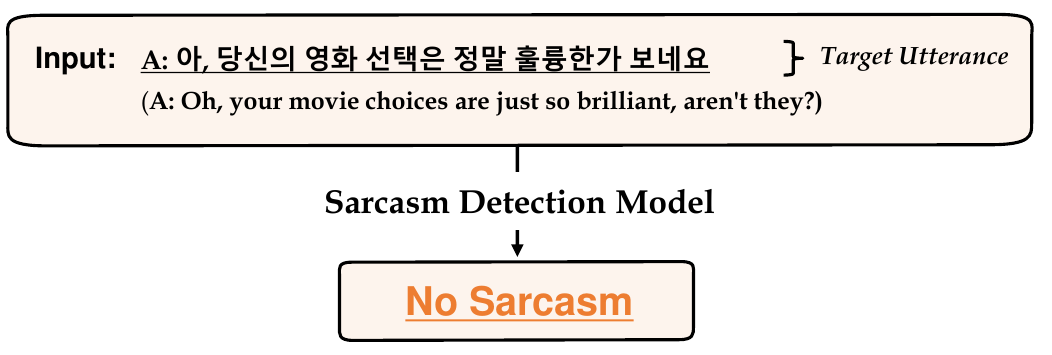}}
    \vspace{-6mm}
\label{fig:intro_fig1_1}}\\ 
    \subfloat[Sarcasm detection w/ context]{\centerline{\includegraphics[width=1\linewidth]{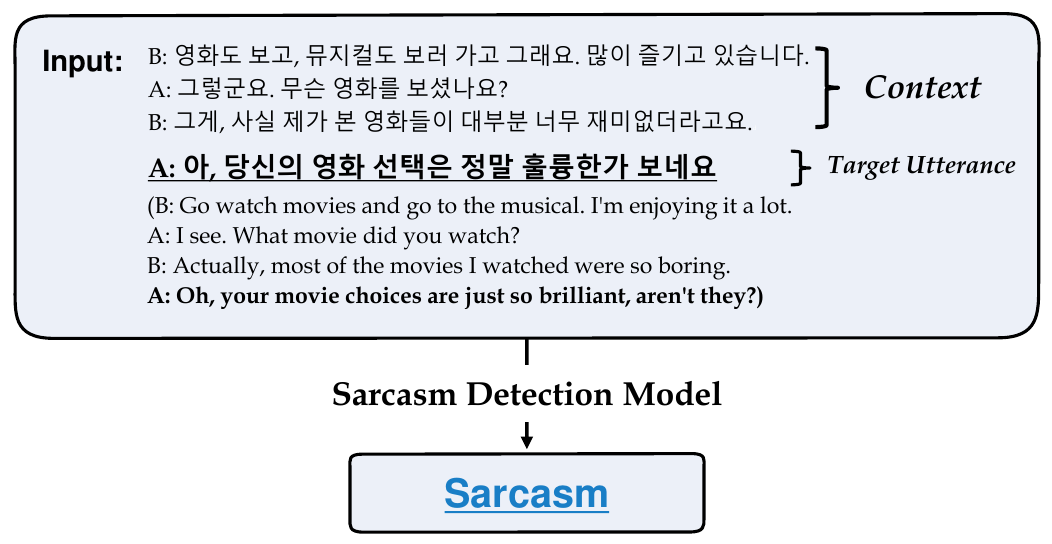}}
\label{fig:intro_fig1_2}}
    \caption{Examples on Korean sarcasm detection results for a target utterance, (a) without the context and (b) with context, respectively.}
    \label{fig:intro_fig}
\end{figure}

Sarcasm detection poses a different challenge compared to general sentiment analysis tasks, primarily due to its sensitivity to the presence or absence of context~\citep{ghosh2017role, avvaru2020detecting}.
For example, as demonstrated in Figure~\ref{fig:intro_fig1_1}, even humans may find it challenging to determine whether the target response in the dialogue is sarcasm when the context is absent.
But when the dialogue context is provided, as illustrated in Figure~\ref{fig:intro_fig1_2}, we can easily understand that the last response is sarcasm used to tease A for not choosing a good movie.
As in this example, context is particularly significant in sarcasm detection tasks~\citep{bamman2015contextualized}. Therefore, it is necessary to develop a context-aware sarcasm detection system that can utilize dialogue history.

Meanwhile, a majority of datasets utilized for training sarcasm detection systems originate from English Twitter and Reddit comments~\citep{moores2022survey}. However, these datasets inherently diverge from daily dialogues that predominantly involve interactions among acquaintances, making it difficult to use the sarcasm detection system trained on these datasets for daily dialogues.
In addition, sarcasm also reflects cultural differences that exhibit the nuances of each country's culture. 
When annotated by annotators from different countries, these differences result in a degree of inconsistency~\citep{liu2014sarcasm, joshi2016cultural}.
Therefore, relying solely on \textit{translationese}\footnote{translations that are unnatural and overly literal.} when investigating sarcasm detection methods in various languages is inappropriate, as the \textit{translationese} may not capture the linguistic nuances of sarcasm. 
As a result, there is an imperative need for monolingual datasets, especially considering that research on English sarcasm detection has been notably more extensive compared to other languages~\citep{rahma2023comprehensive}. 

In this paper, we introduce a new Korean context-aware sarcasm detection dataset composed of 12.8k daily dialogues. We construct the dataset through a comprehensive and effective data generation pipeline facilitated by both LLMs and human revision.
We utilize two kinds of Korean datasets, which comprise daily Korean dialogues and message exchanges, adequately representing the social and cultural background. 
Using these source datasets, we first extract situations from the dialogues and identify the level of politeness of each utterance. By integrating this contextual information into LLMs as prompts, we generate new dialogues that include both sarcastic responses and corresponding explanations. 
Then, we filter harmful and offensive content from the dataset. Finally, expert annotators review the correctness of labels and remove abnormal dialogues to improve the dataset's overall quality further.

We conduct a series of experiments with our dataset to benchmark various baseline systems, including the strong baseline systems trained with our dataset and several Korean LLMs, for the Korean context-aware sarcasm detection task.
Notably, the pre-trained Korean language model KLUE-RoBERTa~\cite{klue2021} fine-tuned with our dataset significantly outperforms GPT-3.5 and achieves performance nearly equivalent to GPT-4~\cite{openai2023gpt4}.
Also, we reveal a significant drop in task performance when context information was omitted, underscoring the importance of considering contextual cues for accurate sarcasm detection. However, it was evident that the model's performance fell behind human capabilities, emphasizing the continued need for research into sarcasm detection methodologies.
The main contributions of the paper can be summarized as follows.
\begin{itemize}
    \item{We propose a comprehensive dataset generation pipeline for the context-aware sarcasm detection task using LLMs and human revision.}
    \item{We introduce a new large-scale Korean Context-aware Sarcasm detection dataset (\textbf{KoCoSa}) through the proposed pipeline, which is composed of 12.8k daily dialogues.}
    \item{We provide a decent analysis of the Korean context-aware sarcasm detection task through this dataset, including the strong baseline system for the task.}
\end{itemize}

\section{Related Work}
\subsection{Sarcasm Detection Dataset}

Numerous English datasets have been released for sarcasm detection~\citep{bamman2015contextualized, rajadesingan2015sarcasm, wallace2015sparse, hazarika2018cascade}, with many of them incorporating context as a crucial factor in determining whether an utterance is sarcastic or not. 
Meanwhile, datasets for non-English languages are less abundant than in English, but some are available. 
Recently, several datasets and research endeavors related to Arabic have been pursued~\citep{farha2020arabic, elgabry2021contextual, faraj2021sarcasmdet, farha2021benchmarking}. Also, few research and datasets for Chinese and Czech have been conducted~\citep{ptavcek2014sarcasm, lin2016sarcasm,
gong2020design, xiang2020ciron}.
To the best of our knowledge, in the case of Korean, \textit{Kocasm} (~\citetlanguageresource{kim2019kocasm}) is the only dataset for sarcasm detection.
However, this dataset lacks the necessary context, highlighting the need for additional Korean datasets. In our work, we first introduce a Korean sarcasm dataset enriched with comprehensive context.

\subsection{Context-aware Sarcasm Detection}

Due to its linguistic nature, many previous studies related to context-aware sarcasm detection have been conducted. Multiple studies have affirmed that the efficacy of sarcasm detection significantly improves with the consideration of context \citep{bamman2015contextualized, ghosh2017magnets, poria2016deeper, ghosh2018sarcasm}. Similarly, \citet{baruah2020context} emphasized that incorporating both the last utterance in the context and the target response proved to be the most effective approach in enhancing sarcasm detection performance. In addition, \citet{dong2020transformer} notes that considering relative context improved sarcasm detection performance. These findings collectively emphasize the crucial role of context-aware approaches in advancing the field of sarcasm detection. 
In this work, we aim to develop a context-aware sarcasm detection system, especially in Korean daily dialogues, by incorporating contextual information.

\begin{figure*}[htb!]
    \centering
    \includegraphics[width=0.95\linewidth]{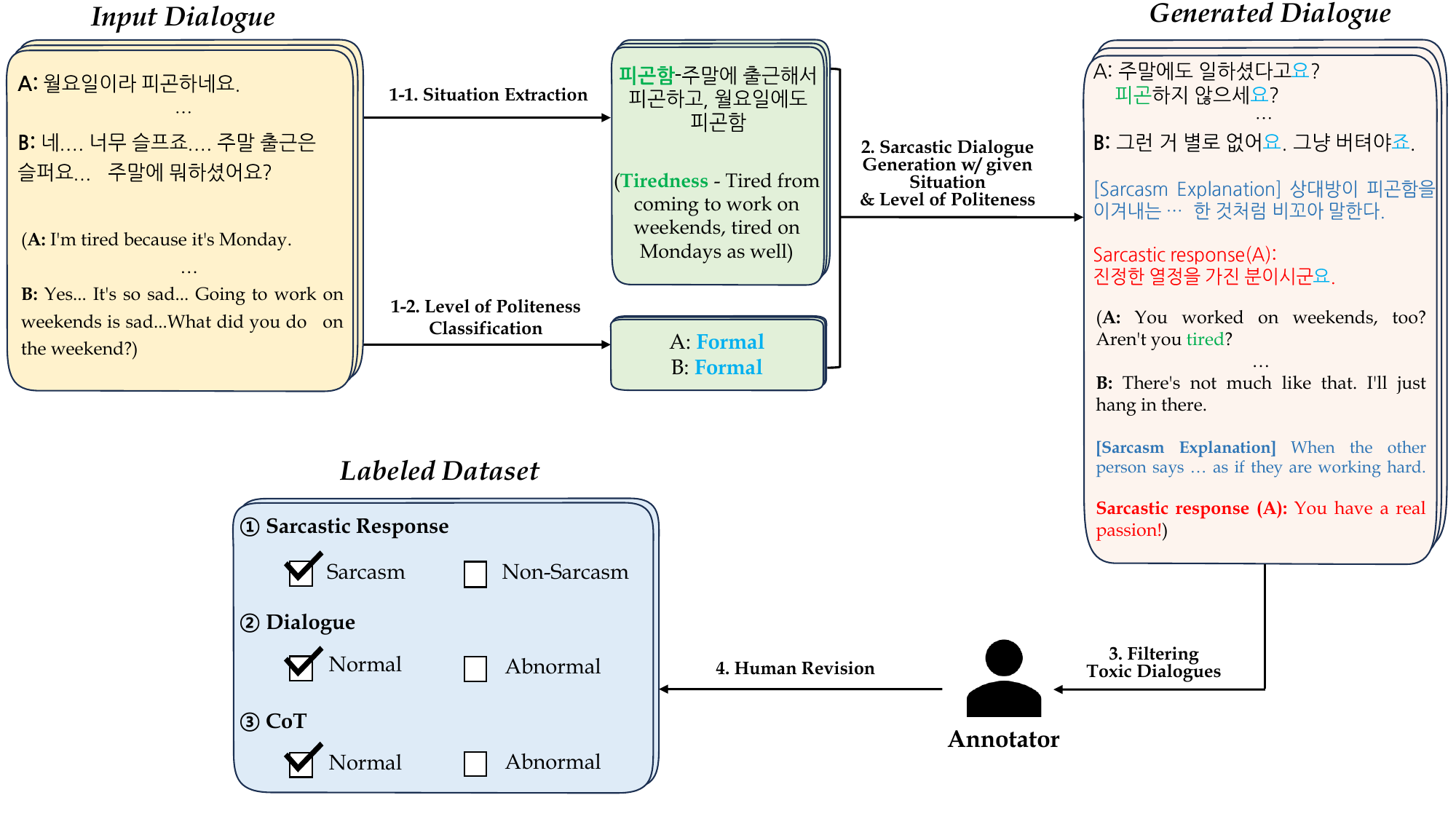}
    \caption{The overall pipeline of KoCoSa dataset construction. In the \textit{Generated Dialogue} example, \textit{light blue letters} represent the honorifics ending of a word in Korean. This figure is best viewed in color.}
    \label{fig:pipeline}

\end{figure*}

\subsection{Data Augmentation Leveraging LLMs}

Previous research has established the efficacy of large pre-trained language models (LLMs) like GPT-3~\citep{brown2020language}, PaLM~\citep{chowdhery2022palm}, and LLaMA~\citep{touvron2023llama} in data augmentation. Similarly, utilizing LLMs for data labeling and annotation has proven to be highly effective in terms of both cost and time efficiency, as demonstrated by~\citep{wang2021want} and~\citep{he2023annollm}. Furthermore, in terms of Zero-Shot augmentation, \citet{kumar2020data} have argued that transformer-based pre-trained language models excel in data augmentation. Recent research has witnessed successful endeavors in data augmentation and labeling by employing LLMs with zero-shot or few-shot prompting~\citep{bonifacio2022inpars, dai2022promptagator, ubani2023zeroshotdataaug}. In this paper, we present a comprehensive pipeline that leverages LLMs for sarcasm detection, providing a holistic approach to data augmentation.

\section{Dataset Construction}
In this section, we describe a comprehensive construction process of our dataset as illustrated in Figure~\ref{fig:pipeline}. Our proposed approach involves harnessing large language models, GPT-4 and GPT-3.5~\cite{openai2023gpt4}, to generate sarcastic dialogues from source dialogues. 
However, we observe that simply instructing LLM to generate sarcastic dialogue has critical problems. 
First, we demonstrate that asking LLM to simply create a new dialogue which includes the sarcastic response for the last utterance, results in sarcastic dialogues with almost similar theme.
Similarly, we find that adding only a sarcastic response at the end of an existing dialogue with LLM often generates an unnatural sarcastic utterance that doesn't match the context. 
Since sarcasm needs to be provided with an appropriate situation, adding sarcastic utterances to a random dialogue mostly creates an abnormal response.
To solve these problems, we propose a new pipeline for generating sarcastic dialogue using LLMs from the existing dialogue as shown in Figure~\ref{fig:pipeline}. 
Rather than simply creating the dialogue from scratch, we construct diverse and natural dialogues that contain sarcastic responses in the last utterance by utilizing the situation extracted from the source dialogue (\S\ref{subsec:generate_dialog}). After automatic filtering on undesired samples (\S\ref{subsec:filtering}), we ask human annotators to review and label these dialogues to construct a high-quality sarcasm detection dataset. (\S\ref{subsec:human_annotation})

\begin{table*}[!t]
{\scriptsize
\centering
\setlength{\tabcolsep}{5pt}
\renewcommand{\arraystretch}{1.2}

\resizebox{2.05\columnwidth}{!}{

\begin{tabular}{p{16cm}}
\toprule
\textbf{Input Prompt} \\
\midrule
You are Korean. You create natural Korean dialogues proficiently. Please consider the level of politeness. \\\textbf{Sarcasm:} someone says something but means the opposite in a mocking or ironic way, often using tone and context to convey the real meaning.\\\textbf{Task Description:} Create a completely new Korean dialogue related to the provided summary. Then, generate a sarcastic sentence in response to the final utterance of the dialogue. Provide an explanation of how to respond sarcastically to the generated dialogue. Then, write a sarcastic response (about 10 to 15 words) without any additional context.\\ \\ \textbf{Example 1.} Situation: 저녁 메뉴-계란 프라이를 태워 먹지 못하는 상황 (\textit{Dinner menu - Couldn't eat because of burnt fried eggs})\\ Level of politeness: A-반말(Informal), B-반말(Informal))
\\ \textbf{A:} 요리는 잘 돼가? (\textit{How's the cooking going?})\\...\\ \textbf{B:} 계란 후라이가 조금 탔어. (\textit{The fried eggs are a little burnt.})\\ \textbf{Sarcasm Explanation:} 계란프라이가 바싹 타버렸다는 마지막 A의 말에 실제로는 부정적인 상황인데, 이 상황을 긍정적인 방향으로 비꼬아 말한다.\\(\textit{It's actually a negative situation when A said that the fried egg was burnt out, but A sarcastically calls this situation in a positive direction.})\\ \textbf{Sarcastic response(A):} 이거 정말 바삭바삭하겠는걸. (\textit{It's going to be really crispy.})\\ \\ \textbf{Example 2.} Situation: \{\textcolor{blue}{situation}\}, \\Level of politeness: A-\{\textcolor{blue}{$p_a$}\}, B-\{\textcolor{blue}{$p_b$}\}\\  
A:\\
\bottomrule
\end{tabular}

}

\caption{Prompt used to generate sarcastic dialogue $D'$ in section~\ref{subsec:generate_dialog}. The words colored \textit{blue} respectively indicates $S$, $p_a$, and $p_b$, extracted from $D$.}
\label{tab:prompt}
}
\end{table*}

\subsection{Collecting Source Dialogue}

We use two primary Korean source dialogue corpora to construct KoCoSa: NIKL Messenger Corpus~\citelanguageresource{nikl2022}(24.4\%) and NIKL Online Text Message Corpus~\citelanguageresource{nikl2022_text}(75.6\%). The data formats of both corpora are nearly identical, with very minor differences. The subject matter of this source dialogue corpora pertains to daily dialogue. 
We find that in cases involving three or more participants, the dialogue lacks consistency due to the unordered nature of utterance sequences in online messengers. Hence, we only use the dialogues that are made between two people for the dataset.  

\subsection{Generating Sarcastic Dialogues}\label{subsec:generate_dialog}
We provide input dialogue $D$ from the source dataset to GPT-3.5 to extract the situation $S$ (\S\ref{subsubsec:situation_extracton}). Then, we ask GPT-4 to generate a new dialogue $D'$ based on $S$ extracted from the previous step to include a sarcastic response $R$ (\S\ref{subsubsec:generating_new_dialogue}). 

\subsubsection{Situation Extraction}
\label{subsubsec:situation_extracton}

We first extract the situation of the given input dialogue $D$, utilizing GPT-3.5 to extract $S$ in a fixed format ($S$ = $\{t;s\}$, by two-shot in-context learning, $t$ and $s$ each representing the main theme and brief summary) as shown in Figure~\ref{fig:pipeline}. 
We observe that if a dialogue is generated from $t$ only, the issue of diversity arises. Similarly, when generating content solely based on $s$, there is a tendency to overly focus on the summary's contents, leading to an increase in abnormalities as the dialogue's subject rapidly shifts. 
For these reasons, we utilize both $t$ and $s$ for a new dialogue generation.
We truncate a total of a maximum of 6 turns as input dialogue $D$ from the source corpus to mitigate multiple distinct situations that may exhibit abrupt transitions in the dialogue.

\begin{table*}[htb!]
{
\scriptsize
\centering
\setlength{\tabcolsep}{5pt}
\renewcommand{\arraystretch}{1.2}
\begin{tabular}{p{2.7cm}|p{10.5cm}|p{1.2cm}}
\toprule
&\textbf{Generated}                                  & \textbf{Label}   \\
\midrule
\textbf{Context} & A: 퇴근이 왜 이렇게 늦어지는 거야? (\textit{Why is leaving work so late?})& \textbf{Normal}\\&B: 너도 늦게 퇴근했나 봐. 나랑 같이 저녁 먹을래? (\textit{I guess you left work late, too. Would you like to have dinner with me?})\\&A: 음, 그래. 뭐 먹을까? (\textit{Well, yeah. What should we eat?})\\&B: 오늘 피곤하니까 그냥 편의점에서 라면 사 와서 끓여먹자. (\textit{I'm tired today, so let's just buy ramen from the convenience store and eat it.})\\
\midrule
\textbf{Response} & A: 그래, 우리 건강에 정말 좋겠다! (\textit{Yeah, it must be great for our health!}) & \textbf{Sarcasm} \\
\midrule
\textbf{Sarcasm Explanation}&B의 마지막 대화에서 건강에 좋지 않은 라면을 추천했기 때문에, A는 이를 아이러니하게 비꼬아 말한다. (\textit{Because 'B' recommended unhealthy ramen in his last utterance, 'A' ironically sarcastically says this.}) & \textbf{Normal} \\
\bottomrule
\end{tabular}
\caption{An example of the generated dataset with annotation results. We deprecate the contexts labeled as abnormal in the released version.}
\label{tab:sample}
}
\end{table*}
\subsubsection{Generating New Dialogue from Situation}
\label{subsubsec:generating_new_dialogue}
We create a new dialogue $D'$ using GPT-4 based on the situation generated from the previous step. As depicted in Table~\ref{tab:prompt}, we configure the sample of the input prompt to create a dialogue with an extracted situation. We fix the level of politeness ($p_a$,$p_b$, each indicating the level of politeness of speaker A and B) of each speaker as \textit{Informal} or \textit{Formal} by examining whether the input dialogue was spoken in honorifics.

\paragraph{Consistent Level of Politeness.}
In Korean, ensuring a consistent level of politeness is of utmost importance. Hence, when a single speaker switches between formal and informal language, it results in a less natural dialogue flow. 
To resolve the issue, we aim to maintain the level of politeness of each speaker in original dialogue by obtaining the formality of the speaker from a pre-trained formal classifier\footnote{\url{https://huggingface.co/j5ng/kcbert-formal-classifier}}. 
We incorporate this formal information into the input prompt to create a dialogue that preserves the formality of speakers in each utterance.

\paragraph{Generating Dialogues with LLMs.}
Using the extracted situation $S$ and level of politeness $p_a$ and $p_b$, we generate a new dialogue $D'$ that contains a sarcastic response in the last utterance using GPT-4. We employ the Chain-of-Thought (CoT) prompting~\cite{wei2022chain} in conjunction with a two-shot generation approach. 
As a way of CoT prompting, we instruct GPT-4 to write \textit{Sarcasm Explanation} that explains why the last utterance would be sarcastic considering the context before creating a sarcastic response. Afterward, a sarcastic response is generated that fits the explanation. 
We present the full prompt in Table~\ref{tab:prompt}.

\subsection{Data Filtering}\label{subsec:filtering}
To improve the quality and moderativity of the dataset, we automatically filter toxic or abnormal dialogues in advance to the annotation process.
\subsubsection{Filtering Toxic Dialogue}
Sarcastic utterances occasionally take the form of offensive language, so it is necessary to filter out data that may cause potential harmfulness when the dataset is disclosed. 
We guarantee the dataset to be moderate through a total of two steps. First, We apply the moderation API of OpenAI\footnote{\url{https://platform.openai.com/docs/guides/moderation}} to all data. Among 17,073 samples, we removed 23(0.0013\%) dialogues that were labeled as "improper."
Then, we additionally filtered five samples that included swear words through the manual data inspection process. For swear word filtering, we use a pre-defined \textit{frequently used toxic words} list.

\subsubsection{Filtering Abnormal Dialogue}
To ensure dataset quality, a final step involves applying automatic abnormal dialogue detection by instructing GPT-3.5 to check if the given dialogue is incongruent.  Any data identified as abnormal through detection is discarded after manual review.


\subsection{Human Annotation}\label{subsec:human_annotation}
We describe the human annotation process for the generated dialogue in this section. We deliberately select exceptional annotators to ensure the attainment of high-quality annotations (\S\ref{subsubsec:annotator_selection}). Furthermore, we furnish comprehensive annotation guidelines, encompassing detailed explanations and a few illustrative samples (\S\ref{subsubsec:annotation_guideline}).

\subsubsection{Annotator Selection}\label{subsubsec:annotator_selection}
We utilize a portion of the data as a preliminary survey for selecting proficient Korean native annotators. In the survey, all participants (30 people) conducted annotations on the same data. We choose 10 annotators by prioritizing those with a high degree (over 65\%) of agreement with the majority vote, thus ensuring the quality of annotations.

\subsubsection{Annotation Guideline}\label{subsubsec:annotation_guideline}
We offer a comprehensive guideline that encompasses a detailed definition of sarcasm, along with relevant examples for various cases to annotators. We provide full dialogue including \textit{Sarcasm Explanation} for the annotation process. We ask annotators to choose three distinct types of labels for each dialogue: \textit{Sarcasm Detection}, \textit{Context Abnormality Detection}, and \textit{Sarcasm Explanation Revision} as follows.
\begin{itemize}
\item{\textbf{Sarcasm Detection} \textit{(Sarcasm / Non-Sarcasm / Abnormal):} Does the last response have sarcastic nuance while congruent with context?
}

\item{\textbf{Context Abnormality Detection} \textit{(Normal / Abnormal):}
Does the provided context seem to be a natural dialogue?
}

\item{\textbf{Sarcasm Explanation Revision} \textit{(Normal / Abnormal):}
Does the explanation offer a suitable description of why the response is sarcasm?}
\end{itemize}



\begin{table}[h!]
{
\scriptsize
\centering
\setlength{\tabcolsep}{5pt}
\renewcommand{\arraystretch}{1.2}
\begin{tabular}{p{1.3cm}|p{1.5cm}|p{1.2cm}|p{1.7cm}}
\toprule
\textbf{Context(C)}&\textbf{Response(R)}& \textbf{Label}&\textbf{Final Format}\\
\midrule
Normal & Sarcasm&\textbf{Sarcasm}&\textbf{C;R}\\
\midrule
Normal & Non-Sarcasm&\textbf{Non-Sarcasm}&\textbf{C;R}\\
\midrule
Normal&Abnormal&\textbf{Non-Sarcasm} &\textbf{C}\\
\midrule
Abnormal&\multicolumn{3}{c}{\textit{Deleted}} \\
\bottomrule
\end{tabular}
\caption{Dataset construction criteria. We delete the dialogue if the label of Context is Abnormal. \textbf{C} denotes Context conversation which consists of a few utterances. \textbf{R} denotes response which is a single utterance. In \textbf{Final Format}, \textbf{C;R} is concatenation of \textbf{C} and \textbf{R}.}
\label{tab:criteria}
}
\end{table}

With the annotated data, we construct the dataset with the construction criteria described in Table~\ref{tab:criteria}. In cases where the context is \textit{Normal} and the response is \textit{Abnormal}, we use the dialogue after removing \textit{Response} from the dialogue.



\begin{table}[h]
\scriptsize
\centering
\setlength{\tabcolsep}{5pt}
\renewcommand{\arraystretch}{1.1}
{%

\resizebox{0.94\columnwidth}{!}{

\begin{tabular}{r|r}
\toprule
Total Dialogues           & 12824 \\ \midrule
Sarcasm & 7608(59.3\%)\\
Non-Sarcasm & 5216(40.7\%)\\
\midrule
Average Turns per Dialogue & 4.3  \\ 
Max Turns                  & 10    \\ 
Min Turns                  & 2     \\ 
Tokens per Dialogue        & 40.3 \\ 
Tokens per Utterance       & 9.3  \\ 
Tokens per Explanation        & 18.9 \\ \bottomrule
\end{tabular}%

}

}
\caption{Overall Statistics of KoCoSa.}
\label{tab:stats}
\end{table}

\begin{table}[hb!]
\scriptsize
\centering
\setlength{\tabcolsep}{3pt} 
\renewcommand{\arraystretch}{1.5} 
\begin{tabular}{p{4.5cm}p{2.3cm}} 
\hline
\textbf{Abnormality Type}  & \textbf{\# of Cases}~(\%)  \\ \hline
Contextual Awkardness      & 788(97.04)                 \\ 
Format Misalignment        & 24(2.96)                   \\ 
\textbf{Total}             & \textbf{812}               \\ \hline
\end{tabular}
\caption{Causes of Abnormal Dialogue}
\vspace{-6mm}
\label{table:abnormal}
\end{table}




\begin{table*}[h]
\scriptsize
\centering

\resizebox{2.05\columnwidth}{!}{

\begin{tabular}{p{0.3\textwidth}ccccccc}
\toprule
\multicolumn{1}{c}{\textbf{Dataset}} & \multicolumn{1}{c}{\textbf{Language}} & \multicolumn{1}{c}{\textbf{Source}} & \multicolumn{1}{c}{\textbf{Sarcastic}} & \multicolumn{1}{c}{\textbf{Total}} & \multicolumn{1}{c}{\textbf{Context}} & \multicolumn{1}{c}{\textbf{Explanation}} \\

\midrule
\citetlanguageresource{Twitter} & English & Twitter & 0.9K & 2.7K & N & N\\
\citetlanguageresource{oraby-etal-2016-creating} & English & Internet Argument & 3.3K & 6.5K & N & N\\
\citetlanguageresource{peled-reichart-2017-sarcasm} & English & Twitter & 3K & 3K & N & N \\
SARC~\citelanguageresource{khodak-etal-2018-large} & English & Reddit & 1.34M & 533M & Y & N \\
Kocasm~\citelanguageresource{kim2019kocasm} & Korean & Twitter & 4.7k & 9.3K & N & N\\
\citetlanguageresource{report-sarcasm} & English & Twitter & 3.1K & 6.2K & Y & N \\
\citetlanguageresource{report-sarcasm} & English & Reddit & 3.4K & 6.8K & Y & N \\
iSarcasm~\citelanguageresource{oprea-magdy-2020-isarcasm} & English & Twitter & 0.8K & 4.5K & N & N\\
\citetlanguageresource{gong2020design} & Chinese & News Comment & 2.5k & 91.8K & Y & N\\
ArSarcasm-v2~\citelanguageresource{abu-farha-etal-2021-overview} & Arabic & Twitter & 3.0K & 15.5K & N & N\\ 
\citetlanguageresource{misra2023Sarcasm} & English & News Headline & 13.6K & 28.6K & N & N\\

\textbf{KoCoSa (Ours)} & \textbf{Korean} & \textbf{Online Message} & \textbf{7.6K} & \textbf{12.8K}& \textbf{Y} & \textbf{Y} \\
\bottomrule
\end{tabular}
}
\caption{Comparison of sarcasm datasets. The \textit{Explanation} column indicates whether an explanation for the sarcasm is included in the data.} 
\label{tab:dataset_comparison}
\end{table*}

\section{Dataset Analysis}
\subsection{Overall Statistics}
\label{subsec:overall_statistics}
KoCoSa dataset contains a total of 12,824 dialogues as shown in Table~\ref{tab:stats}. For the last-utterances that are used for the sarcasm detection task, 7,608(59.3\%) are labeled as \textit{sarcasm}, and 5,216 dialogues(40.7\%) are \textit{Non-sarcasm}. The average number of utterances in the dialogue varies from a minimum of 2 to a maximum of 10 and an average of 4.3.
The initial number of data before the human annotation process is 14,788, and the final dataset consists of 12,824, representing only 13.28\% data loss due to the abnormality of dialogues. This demonstrates the high production efficiency of our dataset construction pipeline. 

Table~\ref{tab:dataset_comparison} compares KoCoSa with other sarcasm detection datasets. KoCoSa represents a pioneering contribution as it incorporates both dialogue context and explanation of sarcasm. Furthermore, KoCoSa distinguishes itself by drawing its source content from online messenger, in contrast to other datasets that primarily derive from online communities like Reddit and Twitter. 

\begin{figure}[]
\centering
    \includegraphics[width=0.95\linewidth]{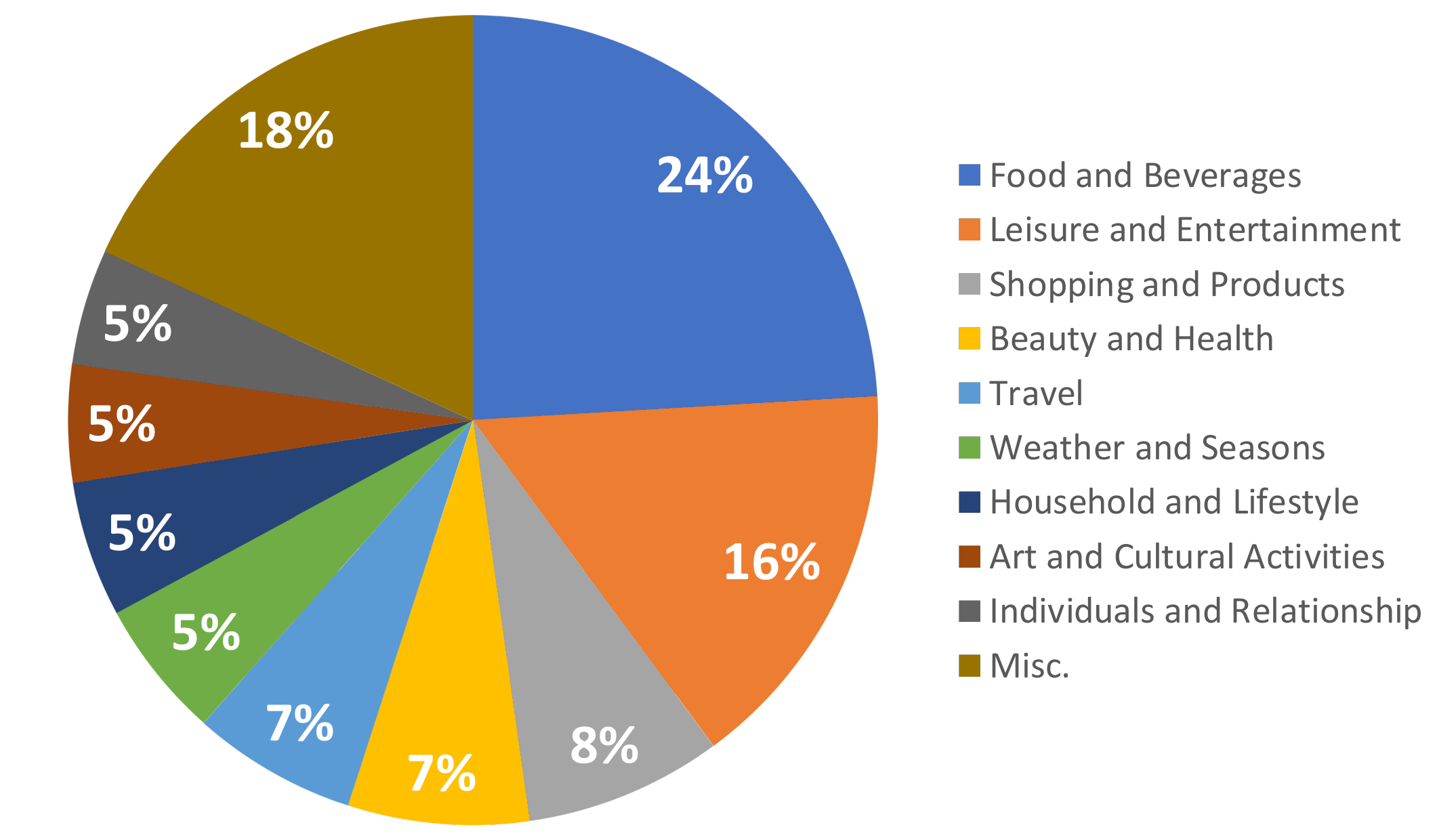}
    \caption{Topic diversity of Online Text Message Corpus and Messenger Corpus. Topics that account for less than 5\% are grouped as Misc.}
    \label{fig:topic}
    \vspace{-5mm}
\end{figure}

We also investigate the topic diversity of dialogues in KoCoSa by analyzing the topics of each dialogue of the source corpora. As shown in Figure~\ref{fig:topic}, source corpora of KoCoSa contains daily dialogues on a variety of topics. Especially, dialogues related to \textit{food and beverage} and \textit{leisure and entertainment} are the top-2 topics accounting for 40\% of the total. Since the dialogues in the KoCoSa dataset are created by extracting the situation from these source corpus, the KoCoSa dataset also contains these diverse topics.



\begin{table}[htb!]
{\scriptsize
\centering
\setlength{\tabcolsep}{5pt}
\renewcommand{\arraystretch}{1.1}{
\begin{tabular}{p{1.95cm}|p{4.9cm}}
\toprule
\textbf{Change Type(\%)} & \multicolumn{1}{c}{$\text{\textbf{Sarcasm}} \longrightarrow  \text{\textbf{Non-Sarcasm}}$} \\
\midrule
\textbf{Casual Dialogue (80\%)} & A: 간식 좀 먹을 게 있어? (\textit{Do we have any snacks?}) \\
& ... \\
& B: 그냥 인터넷으로 주문하면 되지 않을까? (\textit{Why not just order online?}) \\
& \textbf{A: 아니야 직접 가서 사는게 훨씬 편하잖아.} (\textit{No, going there in person is much more convenient.}) \\
\midrule
\textbf{Direct Criticism (20\%)} & A: 이번 주말에 뭐해? (\textit{What are you up to this weekend?}) \\
& ... \\
& B: 그냥 200만 원 정도 생각하고 있어. (\textit{I'm thinking of around 2 million won, just casually.}) \\
& \textbf{A: 와우, 허세 정말 한가득이네.}~\textit{(Wow, that's quite extravagant.)} \\
\bottomrule

\end{tabular}
}
\caption{Sarcasm attempted but labeled as Non-Sarcasm Cases. The ratios next to each type represent the categorization proportions for the random 50 \textit{Non-Sarcasm} dialogues.}
\vspace{-5mm}
\label{tab:label_change_sample}
}
\end{table}
\subsection{Human Quality Evaluation}
\label{sec:Human_eval}

Although our intention is to create natural dialogues that contain sarcastic responses in the dataset, the resulting dataset unexpectedly contains instances labeled as \textit{Non-sarcasm} and \textit{Abnormal}. We conduct a comprehensive analysis of each case and categorize them accordingly.

\subsubsection{Label Change}
\label{subsec:label_change}
We aim to generate dialogues to include sarcastic responses with the proposed pipeline. However, after the human annotation process, we encounter a substantial amount of instances labeled as \textit{Non-Sarcasm}. We systematically categorize them into two distinct types: \textit{Casual Dialogue}, representing regular dialogues with typical responses, and \textit{Direct Criticism}, denoting openly critical statements without any attempt to hide their intent. Among these instances, we notice that \textit{Direct Criticism} is notably shaped by particular word choices. For instance, in the example provided in Table~\ref{tab:label_change_sample}, the word `\textit{extravagant}' straightforwardly conveys a critical tone, resulting in its labeling as Non-Sarcasm.


\begin{table*}[ht!]
\small
\centering
\resizebox{2.07\columnwidth}{!}{
\begin{tabular}{p{0.18\textwidth}ccccccc}

\toprule
\multicolumn{1}{c}{\textbf{Model}} & \multicolumn{1}{c}{\textbf{BA}} & \multicolumn{1}{c}{\textbf{M-F1}} & \multicolumn{1}{c}{\textbf{W-F1}} & \multicolumn{1}{c}{\textbf{Precision-S}} & \multicolumn{1}{c}{\textbf{Recall-S}} & \multicolumn{1}{c}{\textbf{Precision-N}} & \multicolumn{1}{c}{\textbf{Recall-N}} \\
\midrule
\textit{\textbf{Zero/Few-shot}} &  &  &  &  &  & \\
GPT-3.5(zero-shot) & 53.5 & 43.4 & 40.6 & 71.2 & 20.5 & 40.2 & 86.5 \\
GPT-3.5(4-shot) & 51.8 & 42.1 & 39.4 & 66.4 & 20.0 & 39.2 & 83.5 \\
GPT-3.5(8-shot) & 49.4 & 32.3 & 27.2 & 57.7 & 6.0 & 37.8 & \textbf{92.9} \\
GPT-4(zero-shot) & 73.2 & 71.7 & 72.6 & \textbf{83.3} & 68.8 & 60.5 & 77.6 \\
GPT-4(4-shot) & 75.0 & 75.1 & 76.6 & 80.5 & 82.3 & 70.2 & 67.7 \\
GPT-4(8-shot) & 74.5 & 73.9 & 75.1 & 81.9 & 76.3 & 65.4 & 72.6 \\
\midrule
\textit{\textbf{Fine-tuning}} &  &  &  &  &  & \\ 
$\text{KLUE-RoBERTa}_{base}$ & 74.0($\pm$0.2) & 74.1($\pm$0.6) & 74.7($\pm$0.2) & 71.5($\pm$0.2) & \textbf{93.4}($\pm$0.5) & \textbf{87.2}($\pm$0.7) & 54.7($\pm$0.7) \\
$\text{KLUE-RoBERTa}_{large}$ & 74.9($\pm$0.3) & 75.1($\pm$1.0) & 75.5($\pm$0.3) & 74.6($\pm$0.3) & 85.0($\pm$0.6) & 80.0($\pm$0.6) & 64.8($\pm$0.6) \\
\midrule
Human Evaluation & \textbf{80.2} & \textbf{80.1} & \textbf{80.3} & 83.0 & 80.5 & 77.1 & 79.9 \\
\bottomrule
\end{tabular}

}

\caption{Sarcasm detection score on the test set of KoCoSa for various models. 
Note that $\text{KLUE-RoBERTa}$ models are fine-tuned for the training set of KoCoSa.
Balanced Accuracy(\textit{BA}), Macro-F1(\textit{M-F1}) and Weighted F1(\textit{W-F1}) are used. 
\textit{Precision-S} and \textit{Recall-S} represent scores for the \textit{Sarcasm} label. \textit{Precision-N} and \textit{Recall-N} show scores for the \textit{Non-Sarcasm} label.}
\label{tab:baseline_score}
\end{table*}

\subsubsection{Abnormal Case Study}
\label{subsec:abnormal_case_study}
We created a new dialogue using LLMs instead of simply modifying the dialogue of the existing dataset to construct KoCoSa. We found that a considerable proportion of these LLM-generated dialogues are labeled as abnormal after the human annotation. Hence, we manually analyzed the types of these abnormal dialogues within a subset of KoCoSa. These abnormal cases can be broadly categorized into two distinct types:
\begin{itemize}[topsep=5pt, partopsep=0pt, parsep=6pt, itemsep=6pt]
\setlength\itemsep{0em} 
    \item \textbf{Contextual Awkardness}: The unnatural choice of words or context either semantically, morphologically, or both.
    \item \textbf{Format Misalignment}: Cases that do not adhere to the dataset format such as the absence of dialogue, usage of English text, or generation of two or more dialogues, etc.

\end{itemize}
As demonstrated in Table~\ref{table:abnormal}, the majority of cases labeled as abnormal are caused by contextual awkwardness. These issues stem from the inherent characteristics of the source dataset, which encompasses a wide range of topics. This diversity poses a significant challenge in conducting topic extraction and identifying a singular, predominant theme. In order to address the issue of topic diversity, we truncate the dialogues in the source dataset into 6 turns. Despite this approach, we discover that certain remnants of the inherent characteristics of the source dataset persisted. We also observe that a small portion of dialogues are abnormal due to format misalignment.

\section{Experiment}

\subsection{Experimental Setup}
\paragraph{Baselines} 
We provide sarcasm detection scores on the KoCoSa dataset using two zero/few-shot models, GPT-3.5 and GPT-4. Additionally, we fine-tune the $\text{KLUE-RoBERTa}_{base}$ and $\text{KLUE-RoBERTa}_{large}$, which is a pre-trained Korean language model on the Korean Language Understanding Evaluation (KLUE) dataset~\cite{klue2021}, to our dataset for building strong baseline systems.
We set the number of epochs as 5 and use a batch size of 16 for fine-tuning both $\text{KLUE-RoBERTa}_{base}$ and $\text{KLUE-RoBERTa}_{large}$. For GPT-3.5/4, we experiment with in-context learning settings by giving 0/4/8 examples.
\paragraph{Data Preparation} 
We split KoCoSa into Train/Dev/Test in the proportions of about 8:1:1. Since most generated data start with speaker \textit{A}, we exchange the names of speakers to mitigate the possible bias. By randomly selecting 50\% of dataset, we switch the name of speaker \textit{A} to speaker \textit{B}, and the name of speaker \textit{B} to speaker \textit{A}. 

The input data passed to the models consists of contexts, and sarcastic responses, where the output is the label for the sarcasm detection task.
We do not use explanation data for developing sarcasm detection systems in our work, leaving it for future work. 

\paragraph{Evaluation Metric} 
We use seven metrics considering the imbalanced label distribution in the dataset; Balanced Accuracy, Macro F1, Weighted F1, Precision-[\textit{Sarcasm}, \textit{Non-Sarcasm}] and Recall-[\textit{Sarcasm}, \textit{Non-Sarcasm}]. 
\begin{table}[hbt!]
\footnotesize 
\centering
\setlength{\tabcolsep}{3pt} 
\renewcommand{\arraystretch}{1.4} 
\begin{tabular}{p{4cm}p{1.5cm}p{1.5cm}ccc} 
\hline
\textbf{Context} & \textbf{Model} & \textbf{Human}  \\ 
\hline
Only Response   & 73.2       & 62.2      \\ 
Last 1 Utterance + Response  & 73.2 &   -    \\ 
Last 2 Utterance + Response  & 74.9 &   -    \\ 
Last 3 Utterance + Response  & 75.8 &   -    \\ 
Full Context      & \textbf{76.0}     & \textbf{80.2}        \\ 
\hline
\end{tabular}
\caption{Balanced accuracy of sarcasm detection among utterance length of the dialogue context using $\text{KLUE-RoBERTa}_{large}$.}
\label{table:utterance_length}
\end{table}

\begin{table}[hbt!]
\scriptsize 
\centering
\setlength{\tabcolsep}{3pt} 
\renewcommand{\arraystretch}{1.5} 
\begin{tabular}{p{3.2cm}p{0.8cm}p{0.8cm}p{0.8cm}p{0.8cm}p{0.8cm}} 
\hline
\textbf{Topic} & \multicolumn{2}{c}{\textbf{Balanced Acc.}} & \multicolumn{2}{c}{\textbf{Weighted F1}} \\ 
\hline
& \textbf{KLUE} & \textbf{GPT} & \textbf{KLUE} & \textbf{GPT} \\ 
\hline
Food and Beverages            &  71.9  & 73.2 &  71.7  & 73.9 \\  
Leisure and Entertainment     &  73.6  & 72.8 &  74.3  & 73.0 \\
Individuals and Relationship  &  76.1  & 76.8 &  76.8  & 77.2 \\  
Beauty and Health             &  73.5  & 74.7 &  74.5  & 76.2 \\  
\hline
\end{tabular}
\caption{Performance across various topics in the KoCoSa test set, using $\text{KLUE-RoBERTa}_{large}$ and GPT-4 4-shot learning.} 
\label{table:topic_inference_performance}
\end{table}
\subsection{Results}
As reported in Table~\ref{tab:baseline_score}, the baseline scores of models are competitive with the results from the human evaluation, although they fall slightly below the human evaluation score. 
GPT-3.5 shows about 50\% on the Balanced Accuracy representing almost random guess and the lowest \textit{Recall-S} in all of zero-shot (20.5\%), 4-shot (20\%), and 8-shot learning (6\%). 
Meanwhile, GPT-4 shows the competitive score to Human Evaluation. We build KoCoSa utilizing GPT-4, raising concerns about it being tailored to fit GPT-4's tendencies. However, human revision reveals that about 40\% are abnormal or non-sarcasm. Furthermore, GPT-4's detection performance still falls short of human capabilities. 

As shown in Table~\ref{tab:baseline_score}, 
fine-tuning models show competitive performance compared to zero/few-shot models despite the huge gap in the model size. In terms of Balanced Accuracy, $\text{KLUE-RoBERTa}_{large}$ with 337M parameters outperforms most of the GPT-3.5 and GPT-4 models except the 4-shot of GPT-4.
Meanwhile, in zero/few-shot models, \textit{Precision-S} is higher than \textit{Precision-N}. Contrarily, the 71.5\% and 74.6\% on  \textit{Precision-S} of each $\text{KLUE-RoBERTa}_{base}$ and $\text{KLUE-RoBERTa}_{large}$ are lower than 87.2\% and 80\% on \textit{Precision-N} of each model.  


\subsection{Context Length Dependency}
To demonstrate the importance of context in the sarcasm detection task, we compare scores based on different lengths of utterance turns within the context. 
Unlike the baseline experiment setup which includes \textit{Full context}, we design an experiment to provide varying amounts of contexts to the model for validating the effectiveness of context length dependency in the sarcasm detection task.
Starting from the setup w/o context (\textit{Only Response}), we increase the amount of context to \textit{Full Context} to train $\text{KLUE-RoBERTa}_{large}$ model and report the results.
As shown in Table~\ref{table:utterance_length}, using the full context obtains the best Balanced Accuracy of 76.0. From the lowest Balanced Accuracy of 73.2 when only the response is used, the performance steadily improves as more context utterances are incorporated. It implies that language models benefit from sufficient contextual information to enhance the accuracy of sarcasm detection. 
\subsection{Topic Dependency}
We evaluate the consistency in the detection performance of each system among the topic of dialogues in the sarcasm detection task, by measuring the performance of $\text{KLUE-RoBERTa}_{large}$ and GPT-4 4-shot model on the KoCoSa test set for each topic. Given the similarity between the topics in the source dialogue and those generated in KoCoSa (\S\ref{subsec:overall_statistics}), we categorize the KoCoSa using the source dialogue topics. 

As depicted in Table~\ref{table:topic_inference_performance}, we suggest that sarcasm detection performance does not vary with the dialogue's topic. Specifically, dialogues centered around \textit{Individuals and Relationship} reveal the highest sarcasm detection performance for both $\text{KLUE-RoBERTa}_{large}$ and GPT-4. 
Meanwhile, dialogues about \textit{Food and Beverages} show the lowest Balanced Accuracy 71.9\% for $\text{KLUE-RoBERTa}_{large}$ and \textit{Leisure and Entertainment} show the lowest Balanced Accuracy 72.8\% from GPT-4. 
Notably, the gaps between the highest and lowest scores are 4.2\%p and 5.1\%p in both Balanced Accuracy and Weighted F1. 
Therefore we demonstrate that there is not much difference in performance between topics. 
\vspace{-2mm}

\section{Conclusion}
In this paper, we propose a new Korean context-aware sarcasm detection dataset KoCoSa composed of 12,824 dialogues. We construct this dataset through a comprehensive pipeline that leverages large language models, automatic dataset filtering, and human annotations. 
Our dataset construction pipeline represents an efficient framework capable of generating diverse conversational content while maintaining minimal loss. 
We also provide baseline performance on our dataset including GPT models, and train strong baseline systems using KLUE-RoBERTa models for the dataset.
We expect that our proposed dataset will serve as a valuable resource for advancing research in Korean sarcasm detection. Furthermore, we believe that the dataset generation pipeline described in this paper can facilitate researchers in exploring language-specific sarcasm detection in low-resource languages.

\section{Ethics Statement}\label{sec:ethics_statement}
We carefully supervise the process of generating KoCoSa so as not to pose any ethical issues. 
We initiated the annotation process after obtaining approval from the Institutional Review Board (IRB). Moreover, we ensured the ethical integrity of our data by leveraging OpenAI's Moderation API. Additional human verification was conducted to account for cases where toxic expressions remained in the filtered dialogues. We removed offensive language and inappropriate content, guided by the feedback provided by annotators. Also, we pay data annotators above the minimum wage, ensuring fair and equitable compensation for their work.

\section*{Acknowledgement}
This research was supported by Institute for Information \& Communications Technology Planning \& Evaluation (IITP) through the Korea government (MSIT) under Grant No. 2021-0-01341 (Artificial Intelligence Graduate School Program (Chung-Ang University)).

\section{Bibliographical References}\label{sec:reference}

\bibliographystyle{lrec-coling2024-natbib}
\bibliography{lrec-coling2024-example}

\section{Language Resource References}
\label{sec:language_resource}
\bibliographystylelanguageresource{lrec-coling2024-natbib}
\bibliographylanguageresource{languageresource}

\clearpage

\appendix

{
\renewcommand{\arraystretch}{1.4}
\begin{table*}[!t]{
\scriptsize
\centering
\resizebox{2.05\columnwidth}{!}{

\begin{tabular}{|c|lc|}
\hline
\multicolumn{1}{|c|}{\textbf{Change Type}} & \multicolumn{2}{l|}{\textbf{Sarcasm --> Non-Sarcasm}} \\ \hline
\multirow{2}{*}{\textbf{Casual Dialogue}} & \multicolumn{1}{l|}{\begin{tabular}[c]{@{}l@{}}A: 간식 좀 먹을 게 있어? (\textit{Do you have any snacks to eat?})\\ B: 아, 간식이 떨어져서 새로 사야 될 것 같아. (\textit{Oh, I'm out of snacks, so I think I need to buy some more.})\\ A: 뭐 좋아하는 간식이 있어? (\textit{Do you have any favorite snacks?})\\ B: 나? 흠.. 땅콩버터 쿠키 좋아해. (\textit{Me? Hmm... I like peanut butter cookies.})\\ A: 그럼 오늘 가서 사야겠네. 어디서 파는지 알아? (\textit{Then we should go buy some today. Do you know where to buy them?})\\ B: 그냥 인터넷으로 주문하면 되지 않을까? (\textit{Why not just order online?})\end{tabular}} & context \\ \cline{2-3} 
 & \multicolumn{1}{l|}{A: 아니야, 직접 가서 사는 게 훨씬 편하잖아. (\textit{No, going there in person is much more convenient})} & Response \\ \hline
\multirow{2}{*}{\textbf{Direct Criticism}} & \multicolumn{1}{l|}{\begin{tabular}[c]{@{}l@{}}A: 이번 주말에 뭐 해? (\textit{What are you doing this weekend?}) \\ B: 아무래도 추석이 다가와서 백화점 가서 상품권 좀 사야겠어.\\ (\textit{I guess I need to go to the department store to buy some gift certificates since Chuseok is coming.})\\ A: 그래? 얼마나 사려고 해? (\textit{Really? How much are you planning to buy?}) \\ B: 그냥 200만원 정도 생각하고 있어. (\textit{I'm thinking about just \$2,000})\\ \end{tabular}} & Context \\ \cline{2-3} 
 & \multicolumn{1}{l|}{A: 와우, 허세 정말 한가득이네. (\textit{Wow, that's quite extravagant})} & Response \\ \hline
\end{tabular}}
}

\caption{Examples of label change data. 
According to Table \ref{tab:criteria}, we included these data in the released version with a Non-Sarcasm label.}
\label{tab:appendix_labelchange}
\end{table*}
}

\begin{table*}[!t]
{\scriptsize
\centering
\setlength{\tabcolsep}{5pt}
\renewcommand{\arraystretch}{1.2}

\resizebox{2.05\columnwidth}{!}{

\begin{tabular}{p{16cm}}
\toprule
\textbf{Input Prompt} \\
\midrule
You are Korean. You create natural Korean dialogues proficiently. Please consider the level of politeness. Level of politeness: A-\{$p_a$\}, B-{$p_b$}. \\\textbf{Sarcasm:} someone says something but means the opposite mockingly or ironically, often using tone and context to convey the real meaning.\\\textbf{Task Description:} Create a completely new Korean dialogue related to the provided summary. Then, generate a sarcastic sentence in response to the final utterance of the dialogue. Provide an explanation of how to respond sarcastically to the generated dialogue. Then, write a sarcastic response (about 10 to 15 words) without any additional context.\\ \\ \textbf{Example 1.} Situation: 저녁 메뉴-계란 프라이를 태워 먹지 못하는 상황 (\textit{Dinner menu - Couldn't eat because of burnt fried eggs})\\ Level of politeness: A-반말(Informal), B-반말(Informal))
\\ Intimacy: 4
\\ \textbf{A:} 요리는 잘 돼가? (\textit{How's the cooking going?})\\\textbf{B:} 응 지금까지는 순항 중이야. 하나만 빼고.(\textit{Yeah, it's been smooth sailing so far. Except for one thing.})\\\textbf{A:} 뭐가 문제야? 잘 안되는게 있어? (\textit{What's the problem? Is something not going well?})\\ \textbf{B:} 계란 후라이가 조금 탔어. (\textit{The fried eggs are a little burnt.})\\ \textbf{Sarcasm Explanation:} 계란프라이가 바싹 타버렸다는 마지막 A의 말에 실제로는 부정적인 상황인데, 이 상황을 긍정적인 방향으로 비꼬아 말한다.\\(\textit{It's actually a negative situation when A said that the fried egg was burnt out, but A sarcastically calls this situation in a positive direction.})\\ \textbf{Sarcastic response(A):} 이거 정말 바삭바삭하겠는걸. (\textit{It's going to be really crispy.})\\ \\\textbf{Example 2.} Situation: 자기계발-퇴근 후 자기계발을 위해 학원에 등록한 상황 (\textit{Self-improvement - Registering for classes after work for self-improvement})\\ Level of politeness: A-존댓말(Formal), B-반말(Informal))
\\ Intimacy: 3
\\ \textbf{A:} 퇴근하고 뭐 하는거 있어요? (\textit{Do you do anything after work?})\\\textbf{B:} 아니 퇴근하면 힘들잖아. 그냥 집에 가서 쉬어야지. (\textit{No, I'm tired after work. I just go home and rest.})\\\textbf{A:} 저는 얼마 전에 영어학원 등록했어요. (\textit{I recently registered for an English class.})\\ \textbf{B:} 아 진짜? 영어 공부 하려고?? 저번 달에는 중국어 공부할거라며? (\textit{Really? To study English?? Last month, you said you were going to study Chinese.})\\ \textbf{A:} 중국어는 너무 어렵더라고요. 그래서 큰 돈주고 영어학원 다시 등록했어요. (\textit{Chinese turned out to be too difficult. So, I ended up registering for an English class, even if it cost a lot.})\\ \textbf{Sarcasm Explanation:} 영어학원에 등록만 하고 가지 않을 것 같은 상대방의 행동을 긍정적인 기부를 하는 것처럼 비꼬아 말한다. (\textit{Sarcastically comments on the other person's action of registering but not attending the English class as if it's a generous donation.})\\ \textbf{Sarcastic response(B):} 학원에 그렇게 기부를 많이 해도 되는거야? (\textit{Is it okay to make such generous donations to the academy?})\\\\
 Situation-\{\textcolor{blue}{situation}\}, \\Level of politeness: A-\{\textcolor{blue}{$p_a$}\}, B-\{\textcolor{blue}{$p_b$}\}\\  
Intimacy: {\textcolor{blue}{intimacy}}\\
Generate Example: \\
\bottomrule
\end{tabular}

}

\caption{Full prompt used to generate sarcastic dialogue.}
\label{tab:full_prompt}
}
\end{table*}

\section{Construction Details}

\subsection{Hyperparameters}
We employ both GPT-4 and GPT-3.5 large language models (LLMs) for generation. For GPT-3.5, which is utilized to extract situations, we configure it with a maximum of 1000 \textit{tokens}, a \textit{temperature} of 1.0, \textit{top\_p} set to 1.0, and \textit{n} set to 1. In the case of GPT-4, which is used for generating sarcastic dialogues, the settings include a maximum of 3000 \textit{tokens}, a \textit{temperature} of 1.1, \textit{top\_p} at 0.8, n set to 1, and a \textit{frequency penalty} of 0.2. The full prompt including the two-shot examples is described in Table \ref{tab:full_prompt}.
\subsection{Annotation Details}
\subsubsection{Annotator Recruitment}
We recruit annotators from the university's online community. All were well-educated and native Koreans. We present the same 20 pieces of initially generated data (complete dialogue including both context and response) to 30 applicants. Annotation qualifications are granted to 10 applicants who show a match rate of 65\% or more with the majority vote.
\subsubsection{Annotator Compensation}
\begin{figure*}[htb!]
\subfloat[Annotation page example 1.]{\centerline{\includegraphics[width=0.95\textwidth]{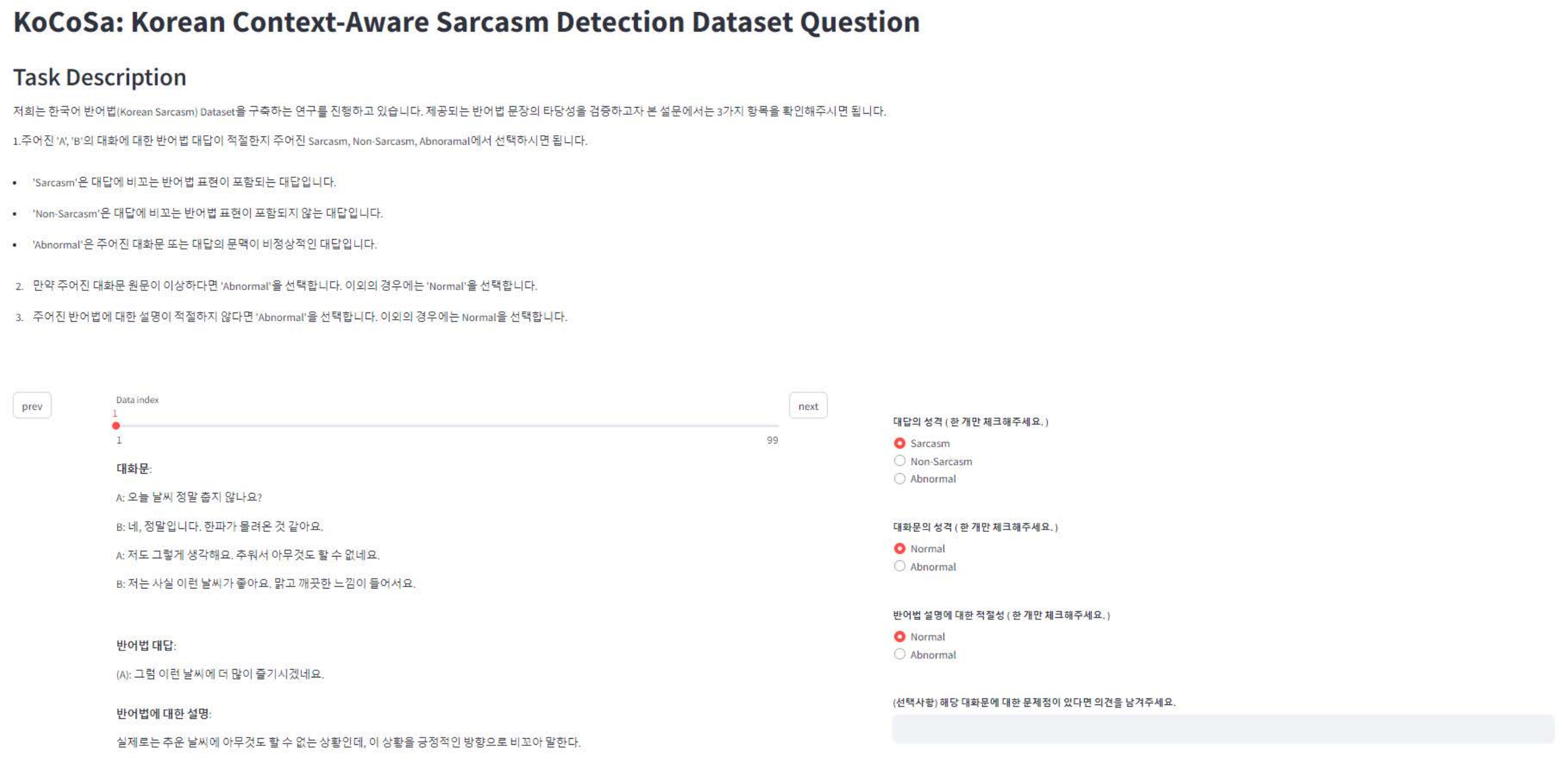}} \label{fig:anno_1}} \hfill
\subfloat[Annotation page example 2.]{\centerline{\includegraphics[width=0.95\linewidth]{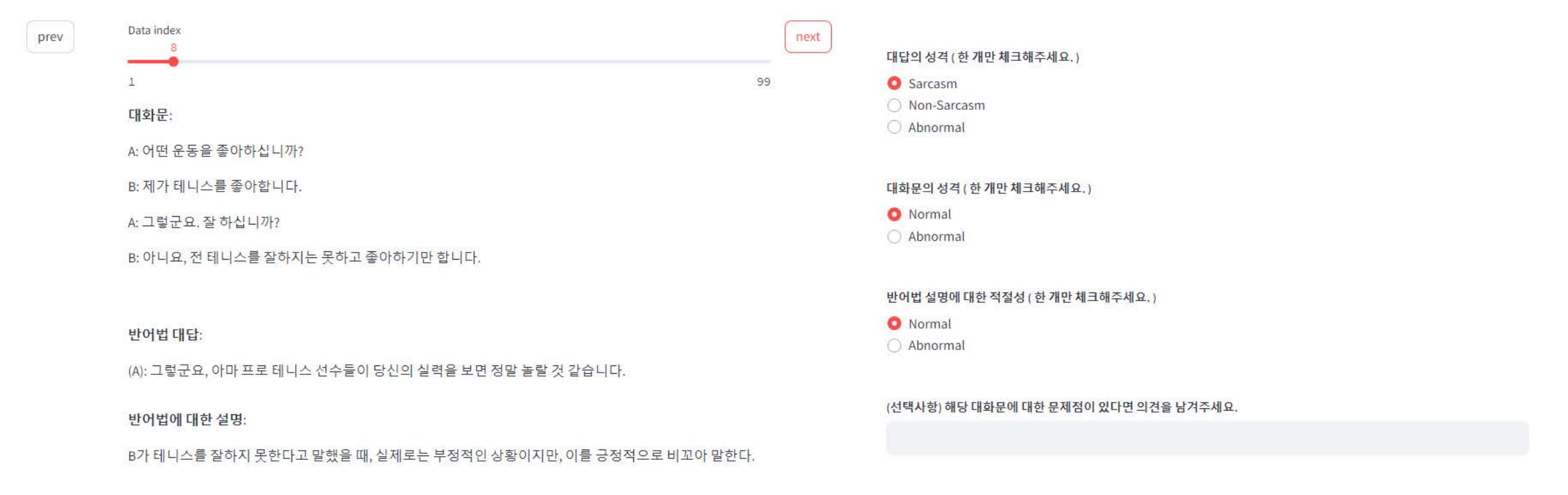}} \label{fig:anno_2} } \hfill
\caption{The task description and data labeling web page for the annotation process. For better understanding, we provide the translation of the \textbf{Task Description} in the figure. \textbf{\textit{Task Description}}: \textit{We are conducting research on constructing a Korean Sarcasm Dataset. In this survey, we would like you to verify the validity of the provided sarcastic sentences by checking the three items below.} The three items refer to the annotation guidelines in section \ref{subsubsec:annotation_guideline}.}
\label{fig:stream_lit}
\end{figure*}

All annotations were conducted in the web page constructed using streamlit\footnote{\url{https://streamlit.io/}}. The total number of samples is 14,788 so each annotator labeled 1,420 to 1,904 samples. Figure \ref{fig:stream_lit} shows the annotation page used in the annotation process. We pay more than \$10/hour, above the minimum wage, based on the hourly workload measured by the preliminary survey.

\begin{table*}[htb!]
\scriptsize
\renewcommand{\arraystretch}{1.8}
\begin{tabularx}{\textwidth}{|>{\centering\arraybackslash\vspace{\fill}}m{4cm}|X|}
\hline
\textbf{Abnormal Context Case}       & \textbf{Example}            \\ \hline
\textbf{Unnatural Choice of Words} & A: 우리 아들, 요즘 왜 그렇게 책을 많이 읽어? (\textit{My son, why do you read so many books these days?})\newline B: \textbf{아들}이 역사에 관심이 많아서 역사책을 많이 읽는다고 해. (\textit{\textbf{My son} is very interested in history, so \textbf{he} reads a lot of history books.})\newline A: 그래? 진짜? 나한테는 아무것도 안 했는데. (\textit{Really? Seriously? He didn't do anything to me.})\newline B: 어제 쇼핑몰에서 할인 쿠폰 찾아서 역사책 사 오더라고. 비싼 책인데도 불구하고. (\textit{Yesterday, \textbf{he} found a discount coupon at the shopping mall and book, despite its high cost...}) \\ \hline
\textbf{Inconsistent Opinion}     & A: 아, 커피 너무 좋아. 아메리카노는 물맛 같고, 라테는 너무 달아서 맛없어. (\textit{Oh, I love coffee. Americano tastes like water and latte is too sweet, so I don’t like them})\newline B: 그래? 나는 아메리카노 좋은데. (\textit{Really? I like Americano.})\newline A: 정말이야? 그럼 나중에 나 커피 사줄 때 아메리카노 사주면 돼. (\textit{Seriously? Then, you can buy me americano when you buy me coffee later.})\newline B: 그래야겠다. 나도 너한테 사줄 때 아메리카노로 사줘야겠네. (\textit{I should. I should buy you an americano when I buy it for you.}) \\ \hline
\textbf{Absence of Dialogue} & Sarcasm explanation: 마지막 B의 말에 따르면, 실제로는 코로나 때문에 콘서트를 가지 못하는데 이것을 긍정적으로 비꼬아 말한다. (\textit{According to B’s last utterance, A is sarcastically making a positive mockery of the fact that B cannot attend concerts because of COVID-19.})\newline Sarcastic response(A): 와, 네 절약 의지 대단하다! (\textit{Wow, your determination to save is impressive!}) \\ \hline    
\end{tabularx}
\caption{Examples of abnormal data in which the context was labeled as abnormal. \textbf{Unnatural Choice of words} and \textbf{Inconsistent Opinion} correspond to \textit{Contextual Akwardness}, while the other (\textbf{Absence of Dialogue}) falls under \textit{Format Misalignment}. As described in Table \ref{tab:criteria}, we have excluded all such data from the released version.}
\label{tab:appendix_abnormal_data}
\end{table*}

\section{Dataset Examples}
As discussed in section \ref{sec:Human_eval}, we aim to create natural dialogues where the final responses are sarcastic. However, LLMs generated unintentional data contrary to our expectations. Here, we show a few examples for the aforementioned data in \ref{sec:Human_eval}.
\subsection{Label Change Data}
\textbf{Label change} refers to the instances where the generated response itself is labeled as non-sarcastic. As we mentioned in section \ref{subsec:abnormal_case_study}, these cases can largely be classified into two categories: Casual Dialogue and Direct Criticism. Examples corresponding to each can be seen in Table \ref{tab:appendix_labelchange}.)

\subsection{Abnormal Context Data}
As described in Table \ref{tab:criteria}, we fully utilized the generated data in various ways. Nevertheless, any instances where the context itself was labeled as abnormal were deleted. Here, we present examples of the removed data in Table \ref{tab:appendix_abnormal_data}.

\section{Experimental Details}
\subsection{Training KoCoSa}
When training $\text{KLUE-RoBERTa}_{base}$ and $\text{KLUE-RoBERTa}_{large}$ on KoCoSa, we set the batch size to 16 and learning rate to 1e-5. 

\subsection{Instruction for GPT-3.5 and GPT-4}
We used the following instruction for GPT-3.5 and GPT-4 detection. 'Task Description: You are really good at detecting the sarcastic response at the last utterance of the given dialogue. If the last utterance is sarcastic, print "1". If not sarcastic, print "0".\verb|\n|Given dialogue: [Input Dialogue]\verb|\n|Detection Result: [Detection Result]'. For 4-shot and 8-shot learning, we provided examples before the input dialogue. Each example is formed as '[Example Dialogue]\verb|\n|Detection Result: [Ground Truth Result]'.  

\end{document}